# PiaNet: A pyramid input augmented convolutional neural network for GGO detection in 3D lung CT scans

Weihua Liu[a], Xiabi Liu[a*], Xiongbiao Luo[b], Murong Wang[a], Guanghui Han[c]，Xinming Zhao[d]，Zheng Zhu[d]

[a]Beijing Lab of Intelligent Information, School of Computer Science, Beijing Institute of Technology, Beijing.
Address: 5 South Zhongguancun Street, Haidian District, Beijing 100081, China
[b]Department of computer science, Xiamen University, Xiamen
Address: No. 422, Siming South Road, Xiamen, Fujian 361005, China
[c]School of Biomedical Engineering, Sun Yat-sen University, Guangzhou.
[d] Dept. of Imaging Diagnosis Chinese, Cancer Hospital, Academy of Medical Sciences, Beijing 100021, China
Address: No.132, Outer Ring Road East, University Town, Panyu District, Guangzhou 510006, China
Email: liuweihua@bit.edu.cn; liuxiabi@bit.edu.cn; xbluo@xmu.edu.cn; wangmurong@bit.edu.cn; zhengguangyuan@bit.edu.cn; hangh3@mail.sysu.edu.cn; xinmingzh@sina.com;dr_zhuzheng@sina.com
Email of Corresponding authors: liuxiabi@bit.edu.cn, xinmingzh@sina.com.

*Abstract*: This paper proposes a new convolutional neural network with multiscale processing for detecting ground-glass opacity (GGO) nodules in 3D computed tomography (CT) images, which is referred to as PiaNet for short. PiaNet consists of a feature-extraction module and a prediction module. The former module is constructed by introducing pyramid multiscale source connections into a contracting-expanding structure. The latter module includes a bounding-box regressor and a classifier that are employed to simultaneously recognize GGO nodules and estimate bounding boxes at multiple scales. To train the proposed PiaNet, a two-stage transfer learning strategy is developed. In the first stage, the feature-extraction module is embedded into a classifier network that is trained on a large data set of GGO and non-GGO patches, which are generated by performing data augmentation from a small number of annotated CT scans. In the second stage, the pretrained feature-extraction module is loaded into PiaNet, and then PiaNet is fine-tuned using the annotated CT scans. We evaluate the proposed PiaNet with the LIDC-IDRI dataset. The experimental results demonstrate that our method outperforms state-of-the-art counterparts, including the Subsolid CAD and Aidence systems and S4ND and GA-SSD methods. PiaNet achieves a sensitivity of 91.75% with only one false positive per scan.

## 1. Introduction

Lung cancer is currently a leading cause of cancer death worldwide, accounting for more than 1.3 million deaths annually (Siegel et al., 2015). Early detection and treatment of lung cancer can improve the survival rate of patients. GGO is a vital sign of lung cancer at its early stage (Henschke et al., 2002), which is defined as increased attenuation of the lung parenchyma without obscuration of the pulmonary vascular markings on the CT images (Jr and Shah, 2005). However, due to their indistinct boundaries, unsharp appearance, and various sizes, GGO nodules are often missed even by experienced radiologists. A promising solution to this problem is the use of computer-aided detection techniques.

Although automatic GGO detection has made significant progress, the previous methods still suffer from the three following problems: 1) they usually assume that the GGO appearance has predefined constraints on CT intensities or spherical shape, which yield unsatisfactory false positive rates, 2) they use hand-crafted visual features, which are sensitive to the high variability in GGO appearance, and 3) they may possibly fail to detect very small nodules or very large nodules.

Recently, deep networks have had an important role in medical healthcare (Uddin and Hassan, 2018; Uddin et al., 2020). In this paper, we propose a new convolutional neural network named PiaNet for accurate GGO detection from 3D CT images. To address the multiscale problem of GGO nodules, we design PiaNet from the idea of multiscale processing. PiaNet consists of a feature-extraction module and a prediction module to simultaneously extract the 3D boxes bounding GGO nodules and compute the corresponding classification confidence scores. We introduce pyramid multiscale source connections into the feature-extraction module to prevent information loss at different scales. Furthermore, the skip connections are applied in the feature-extraction module to preserve the accurate location and multiscale information of detected objects, and the multiscale feature maps are employed in the prediction module to adapt to various sizes of detected objects. To train our PiaNet with a small amount of annotated GGO data, we design a two-stage transfer learning method. Note that we cannot obtain a mass of annotated medical images for training deep networks, which is common in medical applications. In the first stage of our training, the feature-extraction module is solely trained by embedding it into a common classifier network and training this classifier network on a large data set of GGO and non-GGO patches, which are generated by performing data augmentation from a small data set of annotated CT scans. In the second stage, the pretrained feature-extraction module is loaded into PiaNet, and the entire network is fine-tuned with a small number of annotated CT scans. Data augmentation and hard negative mining are adopted in this stage to improve the generalization of training. To evaluate the effectiveness of our proposed PiaNet for GGO detection, we conduct GGO detection experiments on the LIDC-IDRI data set (3Rd et al., 2011), which is the largest publicly available and most utilized data set of lung nodules.

Our main contributions are summarized as follows:

1) We present PiaNet, a deep convolutional neural network for GGO detection that is characterized by a contracting pathway with pyramid multiscale source connections, an expanding pathway with skip connections, and a prediction pathway with multiscale outputs for simultaneous computation of nodule location and classification. This network architecture is novel and demonstrates its value and rationality in ablation studies and comparison experiments. Although the skip connection and multiscale output are not our innovation, the whole architecture of this network can be regarded as our novelty.

2) We firstly introduce multiscale source connections in pyramid form into the neural network. In this way, we can not only extract subtle locations of objects of interest, such as GGOs, but also address the problem of various sizes of objects. As shown in our experiments, this strategy enables consistent gains of performance to different types of networks, including our PiaNet, SSD (Liu et al., 2016) and U-Net (Ronneberger et al., 2015).

3) We design a two-stage transfer learning method to train our PiaNet using a small number of annotated CT scans. Different from commonly employed transfer learning strategies that transfer the knowledge from other domains into the medical domain, we transfer the knowledge between different tasks in the same medical domain, specially, between the discrimination of GGO/non-GGO regions from CT images and the detection of GGO regions from CT images. This two-stage transfer learning method leads to the outstanding performance of GGO detection on the LIDC-IDRI data set. The resultant detection sensitivity is 91.75% at the rate of one false positive per scan, which to the best of our knowledge, is currently the best sensitivity.

The remainder of this paper is organized as follows. Section 2 reviews the related works. Section 3 and 4 presents our PiaNet architecture and its training method, respectively. Section 5 describes the implementation details and discusses experimental results. Section 6 provides some concluding remarks.

## 2. Related Work

### 2.1 Traditional GGO detection approaches

Traditional approaches to computer-aided GGO detection typically consist of two stages (Linying et al., 2017): GGO candidate detection and false positive reduction, where hand-crafted image features are usually used. Bastawrous et al. (Bastawrous et al., 2005) applied a Gabor filter to choose candidate GGOs and used an artificial neural network to reduce false positives. Zhou et al. (Zhou et al., 2006) used cylinder filters suppressing vessels and noises to isolate GGO candidates and classified GGOs by boosting $k$-NN classifiers. Hyoungseop et al. (Kim et al., 2007) extracted tentative regions using the binarization technique and classified the GGO nodules with a linear discriminant function. Jacobs et al. (Jacobs et al., 2011) first used intensity, shape, and texture features to describe candidates' appearance. Subsequently, they applied a linear discriminant classifier and a gentle boost classifier to classify candidate regions. They further introduced a novel set of contexture features and combined it with intensity, shape, and texture features for improving the classification performance (Jacobs et al., 2014).

### 2.2 Nodule detection based on deep learning

In recent years, nodule detection methods based on deep neural networks have achieved state-of-the-art detection performance, including GGO detection. Ginneken et al. (Van Ginneken et al., 2015) presented promising results of nodule extraction using an off-the-shelf convolutional neural network (CNN) that was pretrained on a natural image recognition task. Setio et al. (Setio et al., 2016) used multiple CNNs to extract discriminative features from the candidates, and these features were used to classify candidates as nodules or non-nodule. Superior performance was achieved in the false positive reduction stage. Roth et al. (Roth et al., 2014) proposed an effective 2.5D representation for lymph node detection by taking slices of the CT scans from the point of interests in 3 orthogonal views. The slices were subsequently combined into a 3-channel image and processed by a deep network. Han et al. (Han et al., 2018) proposed hybrid resampling in multi-CNN models for detecting 3D GGO nodules, which reduced the risk of missing small or large GGO nodules.

### 2.3 End-to-end network for object detection

The deep networks have achieved top accuracies on many challenging detection benchmarks of 2D natural images. They can be divided into two categories: two-stage methods and one-stage methods. Two-stage methods generate candidates of bounding boxes and then classify candidates, such as Faster R-CNN (Ren et al., 2017) and Mask R-CNN (He et al., 2017). One-stage methods combine the candidate generation and classification as a whole to become the end-to-end networks, such as Yolo (Redmon et al., 2016) and SSD (Liu et al., 2016). More recently, the end-to-end networks have also been introduced to detect lung nodules in 3D CT images. Similar to Yolo method, Khosravan et al. (Khosravan, 2018) designed a single-scale 3D deep network architecture working in a single shot. Ma et al. (Ma et al., 2018) modified the SSD method and proposed GA-SSD that leverages the attention module with group convolutions for lung nodule detection.

Our PiaNet belongs to the category of one-stage methods. Its structure is U shaped, consisting of an encoder and a decoder.

Differently, Yolo and SSD don't include the decoder. U-shape architectures can also be found in U-Net (Ronneberger et al., 2015) for medical image segmentation, recombinator networks for facial keypoint localization (Honari et al., 2016), feature pyramid networks(FPN) (Lin et al., 2016) and RetinaNet (Lin et al., 2017) for common object detection. The main difference between our PiaNet and other previous U-shape structures exists in the introduction of multiscale source connections and multiscale outputs.

**2.4 Multi-task learning for medical images**

Multi-task learning is helpful to improve the performance of medical image analysis algorithms. Hussein et al. (Hussein et al., 2017) proposed a 3D CNN with a graph regularized sparse multi-task learning for risk stratification of lung nodules. Jie et al. (Jie et al., 2015) discussed a manifold regularized multi-task feature learning approach for multimodality disease classification to preserve both the intrinsic relatedness among multiple modalities of data and the data distribution information in each modality. In our paper, the GGO detection task is divided into two interdependent subtasks: region location and region classification. We try to simultaneously extract the 3D boxes bounding GGO nodules and compute the corresponding classification confidence scores.

## 3. PiaNet Architecture

Due to GPU memory limitations, the input of the PiaNet-based GGO nodule detection method is a CT cube with a fixed size (128×128×128 in our experiments). We preprocess CT scans to meet this demand. Given a 3D CT volume, its voxel size is normalized to 1×1×1 mm$^3$ using bilinear interpolation; then, the CT data in the Hounsfield unit are transformed to [0, 255] by intensity normalization. GGOs must occur in the lung parenchyma, so lung parenchyma segmentation is performed. We divide the resultant volume into 3D cubes in a sliding-window way. Each cube will be inputted into our PiaNet to detect the GGOs in the cube. The final detection result is the combination of all the detected GGOs in each cube.

Fig.1 illustrates the architecture of our PiaNet, where a 128×128×128 CT cube is inputted, and the consequent 47,616 (this number will be explained in this section) bounding boxes are generated and classified as GGOs or non-GGOs. As shown in Fig. 1, which contains three pathways: (i) a contracting pathway with source connections for injecting the multiscale source information for preventing source information loss at different scales; (ii) an expanding pathway with skip connections for combining low-level features and high-level semantics to preserve accurate location information and multiscale information of detected objects; and (iii) a prediction pathway that consists of a regressor and a classifier for making decisions on bounding boxes, where multiscale feature maps are utilized for adapting to various sizes of objects. The contracting and expanding pathways are organized into the feature-extraction module, and the prediction pathway forms the multi-task prediction module.

The details of PiaNet are given in the following subsections.

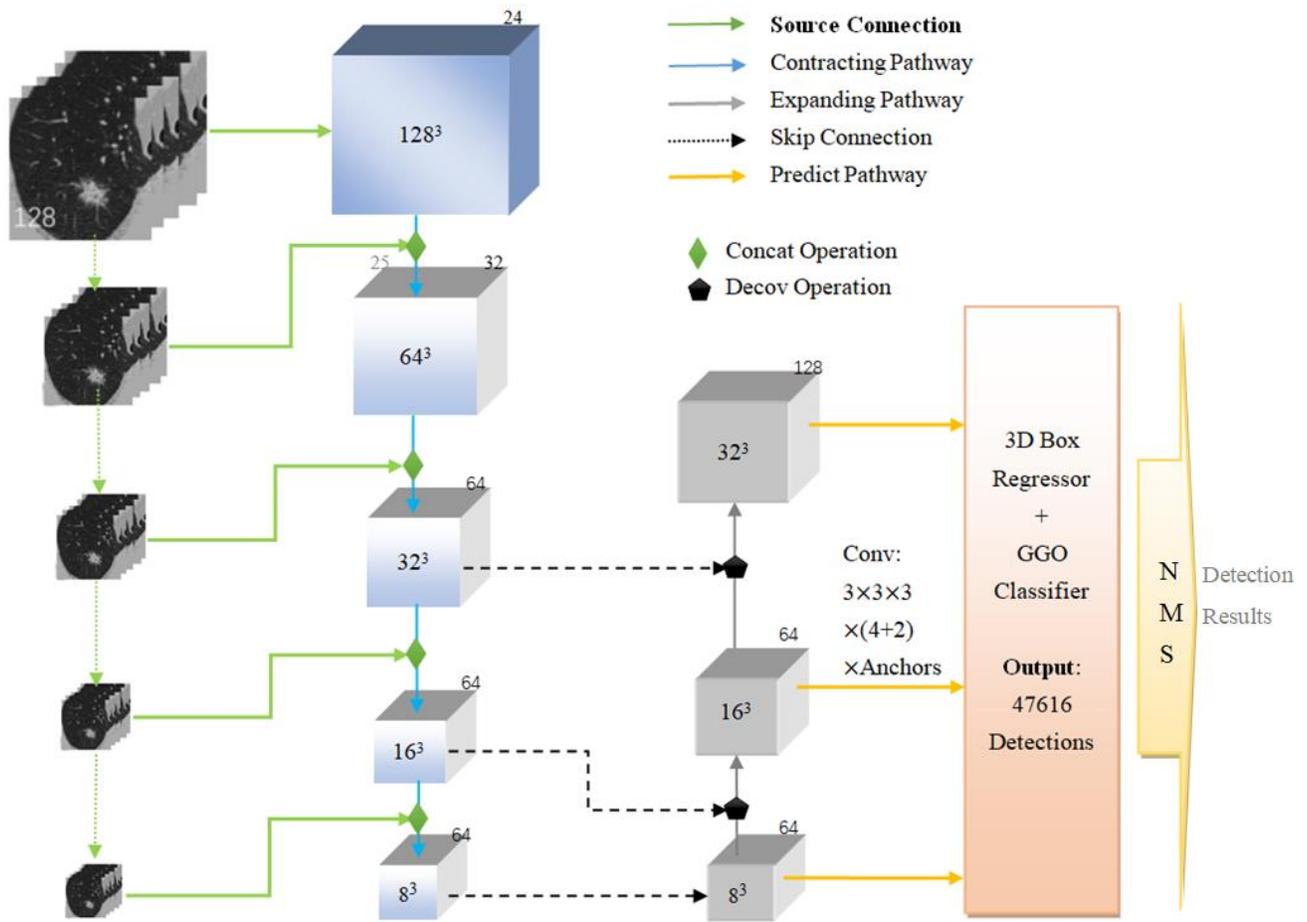

Figure 1. PiaNet architecture (digits on the cube denote the corresponding number of channels)

### 3.1 Contracting pathway with source connections

The contracting pathway is utilized to compute a feature hierarchy that consists of feature maps at several scales with a scaling step of 2. The pathway is composed of computation blocks (simplified as Conv blocks), each of which includes four operations of 3D convolution (Conv), batch normalization (BN), rectified linear units (ReLU), and max pooling (Max).

To capture the spatial information of GGOs at different scales, we construct a multiscale input structure by downsampling the source input with average pooling (Avg) and feeding it into each layer in the contracting pathway, as indicated by the green lines in Fig. 1. These connections are referred to as source connections, with which we concatenate the Avg-pooled source image and feature map of the preceding block as the input of the next block. The green diamond icon in Fig. 1 shows this concatenation operation, which is simplified as "Concat operation". Since source connections involve only one gray value at each spatial position, the block's width is increased only by one (e.g., from 24 to 25, as shown in the first block in Fig. 1).

The integrated computation of the two factors, i.e., computation block and source connection, is illustrated in Fig. 2, where the computation at the largest scale is taken as an example; the processing is the same for other scales.

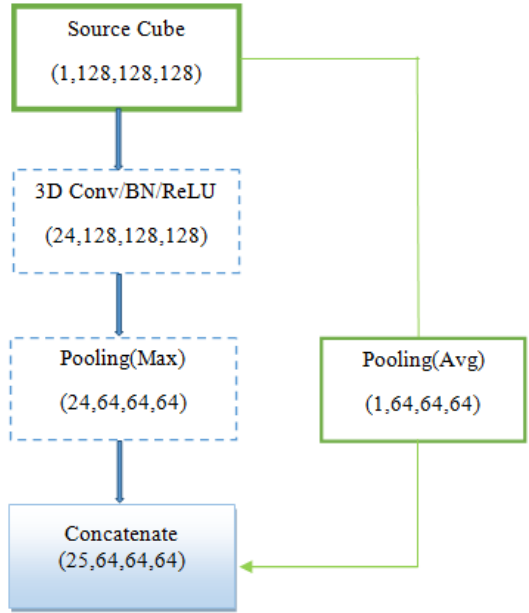

Figure 2. Illustration of the concatenation computation of a computation block and the corresponding source connection, taking the first block of the contracting pathway in Fig. 1 as an example.

The average pooling of source images at various scales produces an image pyramid, as shown in Fig. 3. Thus by adding the source connections to the contracting pathway, the image pyramid can be blended into the network to achieve the ability of multiscale processing with a low-cost computation.

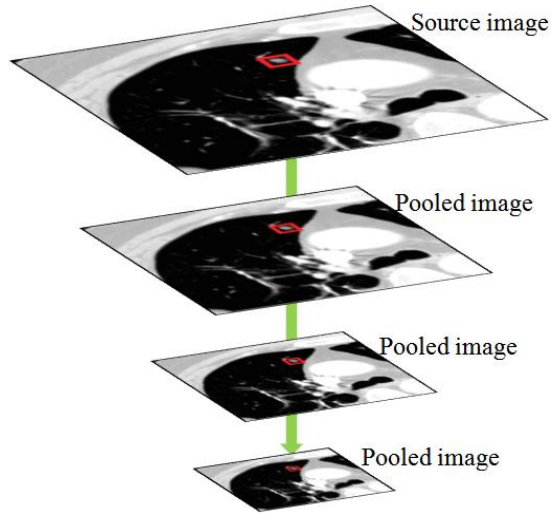

Figure 3. Image pyramid that corresponds to our source connections

### 3.2 Expanding pathway with skip connections

The expanding pathway performs upsampling via Unpooling, Deconvolution, BN, and ReLU operations. Skip connections (Bishop and Bishop, 1995) are further applied to concatenate the feature maps from the contracting pathway with the output at the corresponding scale of the expanding pathway. In this way, we can extract more subtle locations of GGOs in the final result.

By integrating an upsampling procedure and a skip connection, we obtain a deconvolution operation (simplified as Decov operation) in the expanding pathway; the detailed structure is illustrated in Fig. 4. We perform a 1×1×1 convolution operation on the concatenation of the upsampled output from Unpooling-Deconvolution-BN-ReLU operations and the skipped result from the corresponding feature map in the contracting pathway to reduce the aliasing effect.

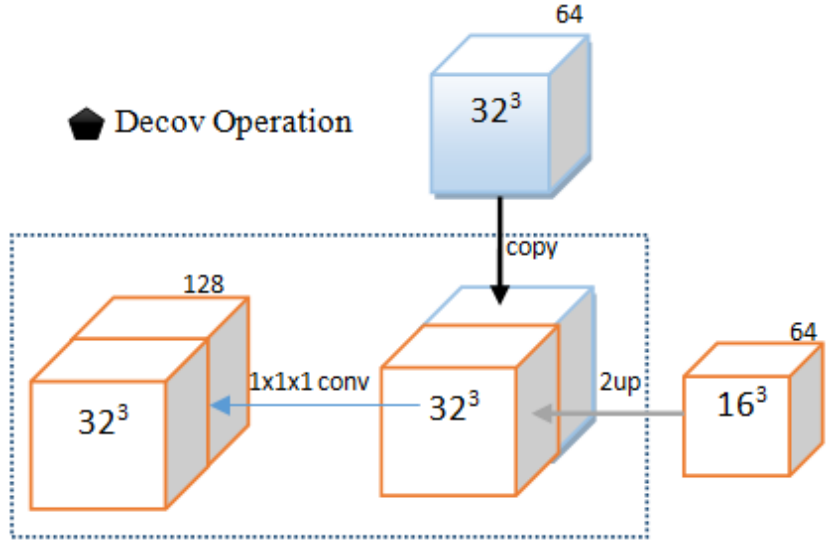

Figure 4. Illustration of the structure of the Decov operation in the expanding pathway, which integrates the upsampled output and skipped information from the contracting pathway.

### 3.3 Prediction pathway

We detect the boxes bounding GGOs and simultaneously compute the corresponding classification confidence scores via the prediction pathway. To achieve this goal, we borrow the anchor box-based idea from Faster R-CNN (Ren et al., 2017), which is introduced as follows.

The CT cube is first partitioned into predefined regular boxes, which are referred to as anchor boxes, that are simplified as anchors in the following descriptions. Each anchor will produce a resultant box with binary classification confidence scores, i.e., we have two types of outputs for each anchor.

To adapt to various sizes of GGOs, we consider multiscale processing that is reflected in two aspects. First, several scales in the expanding pathway are inputted into the prediction pathway and separately processed. For example, the blocks' outputs with scales of 8, 16, and 32 in the expanding pathway shown in Fig. 1 will be processed separately and independently in the following prediction pathway. Second, multisize anchors are considered in the processing for each scale of feature maps, i.e., various sizes of anchors will be considered at each feature map's spatial location.

The structure of a prediction pathway for one scale of feature map (for example, 16) is illustrated in Fig. 5, where "Anchors' Sizes = 3" denotes that 3 different sizes of anchors are considered for this feature map. Since the size of this feature map is $16^3$, $16^3 \times 3$ anchors will be generated, based on which the same number of resultant boxes with classification confidence scores will be produced. The resultant box is a 3D square and is represented by a 4D vector $(x, y, z, r)$, where $(x, y, z)$ is the 3D central point and $r$ is the side length of the square. The corresponding classification confidence scores for GGO and non-GGO form a 2D vector. The convolution computation is performed on the feature map to obtain bounding boxes and classification scores for each anchor. Therefore, we need $3 \times 6$ convolutional kernels, in which $3 \times 4$ convolutional kernels are needed for bounding boxes and $3 \times 2$ convolutional kernels are needed for classification scores, as shown in the left part of Fig. 5 and the right part of Fig. 5, respectively. The shape of each convolutional kernel is $3^3 \times 64$, where $3^3$ is the size of the kernel, and 64 is the number of channels in the feature map. Consequently, we will obtain $16^3 \times 3 = 12,288$ 6D detection results for this scale of the feature map. As previously described, we perform the computation on various scales of feature maps, and the scales of 8, 16, and 32 are considered in our experiments. The number of anchor sizes in the feature maps with scales of 8, 16, and 32 are set to 5, 3, and 1, respectively. Thus, the number of total detection results for an inputted CT cube is $47,616 = 8^3 \times 5 + 16^3 \times 3 + 32^3 \times 1$ in this paper's experiments, as shown in Fig. 1.

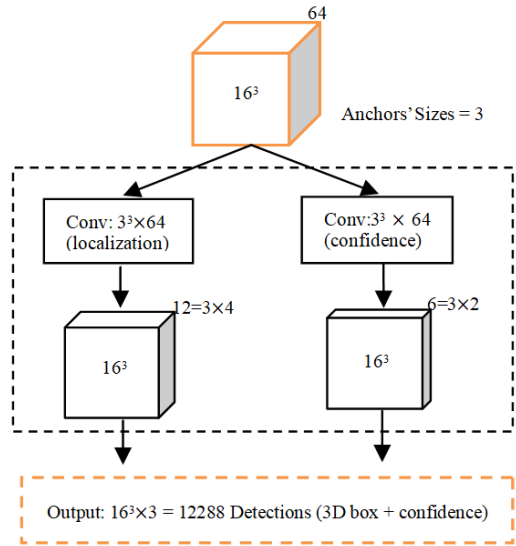

Figure 5. Illustration of the prediction pathway for one scale of the feature map, i.e., the structure of "3D box regressor + GGO classifier" block shown in Fig. 1.

After the resultant boxes and corresponding confidence scores are obtained from the prediction pathway, we perform nonmaximum suppression (NMS) to rule out the overlapping boxes. Among the overlapping boxes, only the box with the maximum confidence score will remain, and other boxes will be deleted. The remaining boxes will be designated GGOs if their classification confidence score is larger than a threshold or designated non-GGO otherwise. This threshold will be established in the applications.

## 4. Training

It is difficult to obtain a mass of annotated training data in medical image applications, which poses a challenge for deep neural network training. This dilemma is also encountered in this work. We can only obtain a small number of 3D CT scans with annotated GGO nodules. To solve the training of our PiaNet with a small amount of data, we develop a two-stage transfer learning method. First, we pretrain our feature-extraction module of PiaNet by embedding it in a GGO classifier network. This network is trained using a large data set of GGO and non-GGO patches that are generated from a small amount of annotated data and by using the data augmentation technique. Therefore, we use a large data set to learn the robust feature-extraction module for discriminating GGO patches from non-GGO patches in the first stage. Second, the pretrained feature-extraction module is integrated into PiaNet, and the whole network is further fine-tuned with a small data set, in which data augmentation and hard negative mining are adopted to improve the generalization of training. This two-stage learning strategy has excellent generalization ability, as demonstrated by our experimental results in Section 5.

We employ the data augmentation technique to transfer the knowledge between different tasks in the same medical domain, specifically, between the discrimination of GGO/non-GGO regions and the detection of GGO regions from CT images. This strategy differs from common transfer learning strategies that transfer knowledge from other domains to the medical domain. Zhou et al. (Zhou et al., 2018) discovered that the gains of transfer learning among the same domain are significantly higher than those from different domains.

In our detection task, we need to locate the positions of GGOs in the image under the guidance of GGO classification and recognize whether the located regions are GGOs. We divide the detection task into these two subtasks and train the network in a multi-task learning way that simultaneously performs position location and region classification. The two subtasks can help each other improve their results. We discover that the features that describe the location and identify whether the region is a GGO are complementary and mutually reinforced in the trained network.

The details of data augmentation and our two-stage training method are discussed in the following section.

### 4.1 Data augmentation

Data augmentation is a feasible solution to solve the problem of small data training by automatically generating more data from the original small data set. We apply the main data augmentation operations to our problem, including flipping, resizing, translating, rotating, and swapping.

1) Flipping: 3D images are randomly flipped with respect to three orthogonal dimensions (coronal, sagittal, and axial).
2) Resizing: 3D images are randomly resized with the ratio in [0.9, 1.1].
3) Translating: 3D images are randomly translated lightly in three axes.
4) Rotating: 3D images are randomly rotated by small degrees.
5) Swapping: 3D images are generated by randomly reordering the values in three axes.

### 4.2 First-stage training

In this stage, the feature-extraction module in PiaNet is joined with an average pooling layer and a two-class softmax layer to construct a GGO classifier for classifying a 3D CT region as GGO or non-GGO. This classifier is trained with a large data set that is generated as follows: First, positive GGO patches and negative non-GGO patches are cropped from a small number of annotated lung CT scans. Second, more positive patches are generated by performing the five operations of data augmentation. Using this large data set of GGO and non-GGO patches, the classifier network is trained by minimizing the softmax-loss function with stochastic gradient descent (SGD). To avoid slow convergence, the weights are initialized with the Xavier initialization method (Glorot and Bengio, 2010). After the training is completed, we obtain the feature-extraction module of our PiaNet. Since the classifier network only involves a binary classification problem and the size of the patches inputted into the classifier network is only 1/8 of the inputted CT cubes of the whole PiaNet, the classifier can be trained quickly to produce an acceptable initial feature-extraction module for more effective training of the whole PiaNet.

### 4.3 Second-stage Training

In this stage, the pretrained feature-extraction module is loaded into PiaNet and further trained with a multi-task learning method. As described in Section 3.3, we attempt to regress from the anchors to the corresponding ground-truth boxes and classify them into GGOs or non-GGOs. These two objectives are combined as a single multi-task objective for training PiaNet.

Formally, let $\Pi_{ij} = \{1, 0\}$ be an indicator that matches the $i$-th anchor (denoted as $a$) with the $j$-th ground truth box (denoted as $g$) in the training data, which is determined according to the intersection over union (IoU) values between the anchors and the ground-truth boxes. Each ground-truth box will be matched with the anchor with its maximum IoU value. Let $c$ be the outputted classification confidence scores, let $l$ be the outputted bounding boxes, and let $g$ be the ground-truth bounding boxes. The multi-task loss function is defined as

$$L(\Pi, c, l, g) = L_{conf}(\Pi, c) + \alpha L_{loc}(\Pi, l, g), \tag{1}$$

where $L_{conf}$ is the loss for the classification confidence, $L_{loc}$ is the loss for the resultant box, and $a$ is the balancing of the contribution of two loss terms and is set to 1 by cross-validation experiments in this paper.

The loss for classification confidence is defined as the binary cross-entropy loss over the confidences of the two classes:

$$L_{conf}(\Pi, c) = -\sum_{i, \Pi_{ij}=1} \log(c_j) - (1-c_j) \sum_{i, \Pi_{ij}=0} \log(1-c_j). \tag{2}$$

Regarding the loss for the resultant box, the modified Smooth $L1$ loss (Girshick, 2015) between $l$ and $g$ is employed. Let $(x, y, z, r)$ be the representation of a box, as described in Section 3.3. Let $t_j^k$ ($k \in \{x, y, z\}$) and $p_j^k$ ($k \in \{x, y, z\}$) be the $k$-th coordinate of center point of the $j$-th ground-truth box and the $j$-th predicted resultant box, respectively, and let $t_j^r$ and $p_j^r$ be the side length of the $j$-th ground-truth box and the $j$-th computed box, respectively. We have

$$L_{loc}(\Pi, l, g) = \sum_{i,j} \Pi_{ij} (\beta \sum_{k \in \{x,y,z\}} Smooth_{L1}(t_j^k - p_j^k) + (1-\beta) Smooth_{L1}(t_j^r - p_j^r)), \tag{3}$$

where Smooth $L1$ function is defined as

$$Smooth_{L1}(x) = \begin{cases} 0.5x^2, & if\ |x| < 1 \\ |x| - 0.5, & otherwise \end{cases}. \tag{4}$$

The weight term $\beta$ is set to 0.6 via careful experiments in this paper, which means that we pay more attention to the center points of the bounding boxes than their side length.

The positions and sizes of both the ground-truth boxes and computed boxes are measured based on the anchors. Let $a_i^k$ ($k \in \{x, y, z\}$) be the $k$-th coordinate of the center point of the $i$-th anchor, and let $a_i^r$ be the side length of the $i$-th anchor. Subsequently, $t_j^k$ ($k \in \{x, y, z\}$), $p_j^k$ ($k \in \{x, y, z\}$), $t_j^r$, and $p_j^r$ in Eq. 3 are computed as

$$t_j^k = (g_j^k - a_i^k) / a_i^r, \tag{5}$$

$$p_j^k = (l_j^k - a_i^k) / a_i^r, \tag{6}$$

$$t_j^r = \log(g_j^r / a_i^r), \tag{7}$$

$$p_j^r = \log(l_j^r / a_i^r), \tag{8}$$

respectively.

Our multi-task learning is performed with training data to minimize Eq. (1) by using the SGD algorithm. The training data in this stage are collected as follows: First, GGO patches are randomly cropped from the annotated lung scans. These patches are augmented by using the flipping and resizing operations described in Section 4.1. The remaining 3 data augmentation operations are also tested here but fail to show significant improvement. Second, we adopt a hard negative mining technique (Shrivastava et al., 2016) to solve the class unbalance problem, in which the number of negative patches is substantially greater than that of positive patches in the detection.

The hard negative mining is performed in three steps. First, all the CT cubes in the training data set are processed by PiaNet to obtain tens of thousands of detections, each of which is associated with a resultant box and a classification confidence score. Second, initial negative examples are randomly selected from these boxes, where the number of selected examples is determined by experience. Third, all the selected negative examples are sorted in descending order of their confidence scores, and the top K examples are chosen as the final negative examples that are employed in training. In this way, the number of negative examples and that of positive examples will be balanced, and hard negative examples will be selected to improve the training.

## 5. Experiments

In this section, we first describe the experimental setup and implementation details of PiaNet. Then we report the results of two groups of experiments. The first group of experiments is conducted to evaluate the roles of PiaNet components, compared with classical networks including U-Net and SSD. The second group of experiments is conducted to test the overall performance of PiaNet and compare it with that of other counterparts, including two state-of-the-art methods and two CAD systems. All experiments were conducted on a computer with four NVIDIA GeForce TITAN Xp GPUs.

### 5.1 Experimental Setup

#### 5.1.1 Dataset

The dataset is collected from the largest publicly available database of lung nodules, the LIDC-IDRI dataset (3Rd et al., 2011). The dataset contains 1018 CT scans and XML-based annotations to stimulate the development of CAD methods for lung nodule detection, classification, and quantitative assessment. Each scan was annotated by 4 experienced thoracic radiologists using a two-phase reading process. Following Naidich et al. (Naidich et al., 2013), we discarded scans with slice thicknesses greater than 3 mm, considering 3D detection in this paper. Scans with inconsistent slice spacing or missing slices can be considered noise and were also excluded. The annotated nodules in the LIDC-IDRI dataset are categorized based on the morphological characteristics and are scored by the radiologists using a 5-point score. Only nodules with scores less than 5 are considered real GGOs and selected for our experiments. Furthermore, we only consider the nodules that are recognized by the majority of the radiologists, at least 3 of 4 radiologists. The GGO nodules that remain after the processing are the targets that the algorithms should detect. Other excluded GGO nodules are labeled "irrelevant findings" and are not selected as true positives or false positives to alleviate the problem of disagreement regarding the definition of a GGO nodule. This strategy is also adopted in the LUNA16 challenge (LUNA16).

A total of 302 CT scans from 299 patients were collected from the LIDC-IDRI data set, in which 635 GGO nodules (271 nodules recognized by 3 radiologists and 364 nodules recognized by 4 radiologists) were included. The diameter of the nodules varies from 3 to 34 mm, with a median of 10.3 mm. We use 250 scans in these 302 scans as the training set and use the remaining 52 scans as the test set.

For one 3D CT scan, we can produce dozens of 128×128×128 cubes. For example, the size of one CT scan is 512×512×206. After preprocessing with voxel size normalization, intensity normalization, and lung parenchyma segmentation, as described in the first paragraph of Section 3, the size of the processed scan is reduced to 252×222×192. This 3D CT volume is divided into 128×128×128 cubes in a sliding-window way, where the stride is set to 64. Consequently, we produced 24 cubes for this CT scan. Note that the roughly labeled lung parenchyma for each scan is mainly provided in the LUNA 16 challenge, which is obtained using an automatic lung segmentation algorithm (Van Rikxoort et al., 2009). For some scans unseen in the LUNA 16 challenge, we manually segmented the lung parenchyma.

#### 5.1.2 Evaluation criteria

In our experiments, we perform a free-response receiver operating characteristic (FROC) analysis (Kundel et al., 2008) to evaluate the performance of GGO detection. The sensitivity and false positives are extensively utilized for evaluating the detection performance. The sensitivity is defined as the fraction of detected true positives divided by the number of all true nodules. The FROC curve reflects the relationship between the sensitivity and the false positives per scan (FPs/scan), in which the sensitivity is plotted as a function of the average number of false positives per scan (FPs/scan). The overall evaluation from the FROC curve is measured as the average of the sensitivities at seven predefined FPs/scan: 1/8, 1/4, 1/2, 1, 2, 4, and 8. This performance metric was introduced into the ANODE09 challenge and referred to as the competition performance metric (CPM) (Niemeijer et al., 2011).

### 5.1.3 Network configuration

Table 1 summarizes the detailed configuration of PiaNet, which is applied in the experiments. As shown in Table 1: 1) in the contracting pathway, five Conv-BN-ReLU-Pooling blocks with source connections are constructed; 2) in the expanding pathway, only three layers are constructed due to the GPU memory limitation. Additionally, due to this issue, we have to control the output size in this pathway. Therefore, although we expect an output shape of (128, 16, 16, 16) when expanding from the scale of 8 to 16, we can only produce the output with the shape of (64, 16, 16, 16) in practice, as shown in Fig. 1; and 3) in the prediction pathway, 3×3×3 convolutions are applied to both the bounding box regressor and the GGO classifier. All the scales of information decoded in the expanding pathway, including 32×32×32, 16×16×16, and 8×8×8, are employed and inputted into the prediction pathway. The number of anchor sizes are set to 1, 3, and 5 for these three scales, which lead to a total of 9 anchor sizes: 4, 6, 8, 10, 12, 16, 20, 26, and 32 mm. Their details are listed in Table 2. In summary, we will generate 47,616 (=$32^3 \times 1 + 16^3 \times 3 + 8^3 \times 5$) anchors for each CT cube, based on which 47,616 resultant boxes and their classification confidence scores will be computed.

Table 1: PiaNet configuration used in the experiments (OP: operation)

|  | Layer Number | Name | Output Shape |
| --- | --- | --- | --- |
| Source | 0 | Input | (1, 128,128,128) |
| Contracting pathway | 1 | Conv block | (24, 64,64,64) |
|  | 2 | Concat OP | (25,64,64,64) |
|  | 3 | Conv block | (32,32,32,32) |
|  | 4 | Concat OP | (33,32,32,32) |
|  | 5 | Conv block | (64,16,16,16) |
|  | 6 | Concat OP | (65,16,16,16) |
|  | 7 | Conv block | (64,8,8,8) |
|  | 8 | Concat OP | (65,8,8,8) |
|  | 9 | Conv block (No pooling) | (64,8,8,8) |
| Expanding pathway | 10 | 1x1x1Conv | (64,8,8,8) |
|  | 11 | Decov OP | (64,16,16,16) |
|  | 12 | Decov OP | (128,32,32,32) |
| Prediction pathway | 13 | BoxRegressor/ Classifier | (4,32,32,32)/ (2,32,32,32) |
|  | 14 | BoxRegressor/ Classifier | (12,16,16,16)/ (6,16,16,16) |
|  | 15 | BoxRegressor/ Classifier | (20,8,8,8)/ (10,8,8,8) |
| Detection | 16 | Output | (47,616,4,1,1)/ (47,616,2,1,1) |

Table2. Configuration of anchor sizes for each scale of decoded feature maps (#:number of)

| Feature map | # anchor size | # anchor box | Size (mm) | | | | |
| --- | --- | --- | --- | --- | --- | --- | --- |
| 32×32×32 | 1 | 32,768 | 4 | \ | \ | \ | \ |
| 16×16×16 | 3 | 12,288 | 6 | 8 | 10 | \ | \ |
| 8×8×8 | 5 | 2,560 | 12 | 16 | 20 | 26 | 32 |
| Sum |  | 47,616 | | | | | |

### 5.2 Experimental Results
#### 5.2.1 Ablation experiments

Different from other network structures for object detection, our network architecture is designed on the basis of the idea of multiscale processing, which is realized in the combination of three strategies. First, pyramid multiscale source connections are introduced into the contracting pathway to prevent source information loss at different scales. Second, the expanding pathway with skip connections is adopted for extracting the accurate location and multiscale information of detected objects. Third, the use of multiscale feature maps in the prediction pathway contributes robustness to various sizes of objects. We conduct controlled experiments to examine how each of these three strategies affects performance. For all the experiments, we use the same training data set and data augmentation operations; the hyperparameters of the network are also shared.

If the source connections are disregarded, then our PiaNet is closely related to SSD (Liu et al., 2016) and U-Net (Ronneberger et al., 2015). SSD is a popular single-shot detector for natural images, which produces predictions from multiscale feature maps but only in the contracting pathway. Conversely, U-Net has an elegant contracting-expanding structure, but it only uses the last feature map in the prediction and disregards the multiscale processing.

To show the importance of introducing the source connections into the networks, we conduct six experiments of GGO detection with or without source connections based on PiaNet, SSD and U-Net. Table 3 lists the results of these experiments, which shows that source connections improve the sensitivity from 78.2% to 83.7% for SSD, from 82.4% to 86.9% for U-Net, and from 87.3% to 94.2% for PiaNet, at a slightly better FPs/scan rate. The maximal boosted performance reaches 6.9 points. The important observation is that the gains with our source connections are consistent among different networks.

Table 3 also shows that our expanding pathway and multiscale prediction has an important role in improving the performance. To understand the importance of the expanding pathway, we can compare the performance of SSD and PiaNet(-). The main difference between them is whether the expanding pathway is utilized. We observe that the sensitivity is improved from 78.2% of SSD without an expanding pathway to 87.3% of PiaNet(-) with an expanding pathway. Note that some methods that have been improved by adding an expanding pathway in the SSD have been proposed, such as the method by Fu et al. (Fu et al., 2017), which has an encoder-decoder structure, similar to PiaNet without source connections, i.e., PiaNet(-). By comparing PiaNet(-) with PiaNet, we discover that our source connections make our network significantly surpass these types of SSD extended methods. For the importance of multiscale prediction, we can compare the performance of U-Net and PiaNet(-) and determine that the sensitivity is improved from 82.4% of U-Net without multiscale prediction to 87.3% of PiaNet(-) with multiscale prediction.

Table 3. GGO detection results using SSD, U-Net, and PiaNet (+: with source connections, -: without source connections)

| Methods | Multiscale feature maps? | Expanding pathway? | Source connection? | Sensitivity | FPs/scan |
|---|---|---|---|---|---|
| (a) SSD | Y | / | N | 78.2% | 9.1 |
| (b) SSD(+) | Y | / | Y | 83.7% | 8.9 |
| (c) U-Net | / | Y | N | 82.4% | 7.8 |
| (d) U-Net(+) | / | Y | Y | 86.9% | 7.7 |
| (e) PiaNet(-) | Y | Y | N | 87.3% | 8.5 |
| (f) PiaNet | Y | Y | Y | **94.2%** | **6.7** |

To further evaluate the effect of using multiscale feature maps for prediction, we progressively remove feature maps and observe the change in performance. Table 4 shows a stable decrease in accuracy, from 94.2% to 90.3%, with a reduction in the feature map scale. The findings show that we need sufficient scales to cover the wide range of GGOs' sizes.

Table 4. Effects of using multiscale feature maps

| Prediction from features maps: | | | Anchor box | Performance | |
|---|---|---|---|---|---|
| $32 \times 32 \times 32$ | $16 \times 16 \times 16$ | $8 \times 8 \times 8$ | #number | Sensitivity | FPs/scan |
| Y | Y | Y | 47,616 | 94.2% | 6.7 |
| Y | Y | \ | 45,056 | 92.5% | 6.8 |
| Y | \ | \ | 32,768 | 90.3% | 7.1 |

Table 3 reveals that our PiaNet achieves the highest sensitivity (94.2%) at the lowest FPs/scan rate (6.7) among these detection experiments, which demonstrates the superiority of our network architecture to U-Net and SSD.

**5.2.2 Comparison experiments**

In this section, we compare the GGO detection performance of our PiaNet with two state-of-the-art methods, GA-SSD (Ma et al., 2018) and S4ND (Khosravan, 2018), and two real CAD systems, SubsolidCAD (Jacobs et al., 2014) and Aidence (Aaa et al., 2016). GA-SSD is an improved method that is based on SSD by implementing the attention mechanism and the group convolution, and the S4ND method is a single-shot and single-scale method. SubsolidCAD is a conventional system that uses 4 categories of hand-crafted features to describe the appearance of GGOs. The system focuses on GGO nodule detection and can reach a sensitivity of 80% at an average rate of one false positive per scan; the corresponding CPM value is 0.734. The Aidence system is based on convolutional networks. It was trained on a subset of studies from the National Lung Screening Trial (White, 2011) and achieved the best CPM of 0.764 in the LUNA16 challenge (Aaa et al., 2016). The LUNA16 challenge data set contains 253 GGO nodules from the LIDC-IDRI dataset. All of the nodules are contained in our experimental data.

Fig. 6 compares the performance of GA-SSD, S4ND, Subsolid CAD, Aidence system, and PiaNet. The findings show that PiaNet outperforms other methods. Our method achieves a CMP of 0.8826, which is better than a CMP of 0.8547 achieved by the GA-SSD, a CMP of 0.8666 achieved by S4ND, a CMP of 0.734 achieved by the SubsolidCAD system, and a CMP of 0.767 achieved by the Aidence system. The corresponding performance increases for GA-SSD, S4ND, SubsolidCAD and Aidence are 3.26%, 1.85%, 20.25% and 15.07%, respectively.

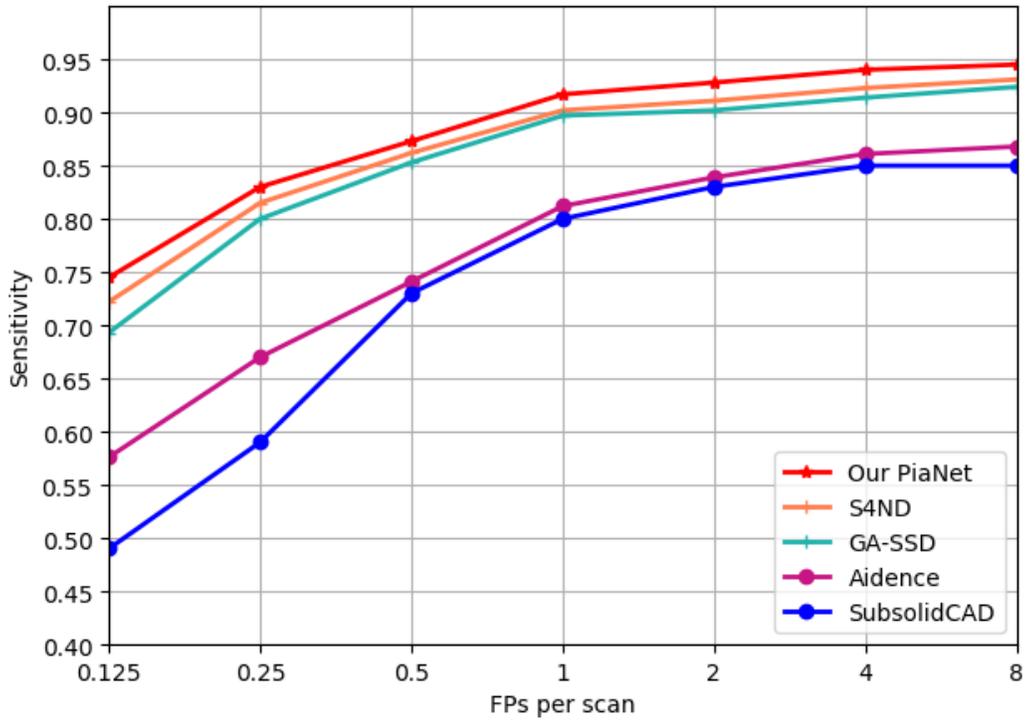

Figure 6. Comparison of performance from our PiaNet, GA-SSD, S4ND, SubsolidCAD and Aidence.

### 5.3 Detection Visualization

In Fig. 7, we visualize the detection results, including the correct results and a few errors, which are produced from PiaNet by establishing the threshold of confidence scores for recognizing GGO to be 0.8. These example results demonstrate that PiaNet is robust and can identify GGO nodules with various sizes (from 3 mm to 33 mm in our test set) and irregular shapes and textures.

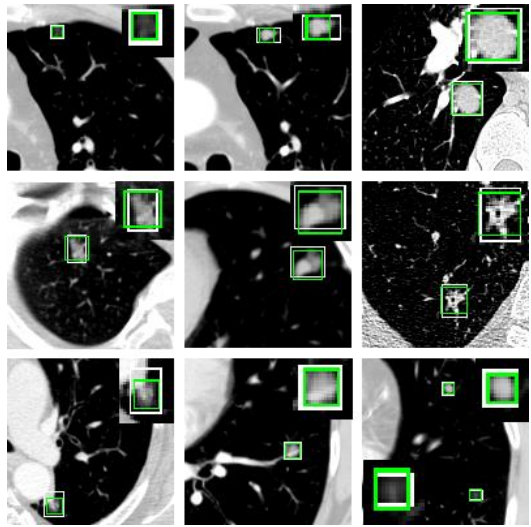

Figure 7. Examples of correct detections by PiaNet on the LIDC-IDRI dataset, where the white rectangles denote the ground-truth boxes; the green rectangles denote the detection results, and the nodules are shown in zoomed-in areas at the top-right corners.

Examples of missed GGO nodules (false negatives) and false-detected results (false positives) by PiaNet are shown in the top row of Fig. 8 and bottom row of Fig. 8, respectively. We can see that the missed nodules are notably low in contrast or very uncommon, and the false-detected nodules are usually located very close to the other tissues with a similar appearance, such as blood and chest wall. To address these hard cases to further reduce the error rate, more discriminative features should be learned by making the network training focus on hard examples.

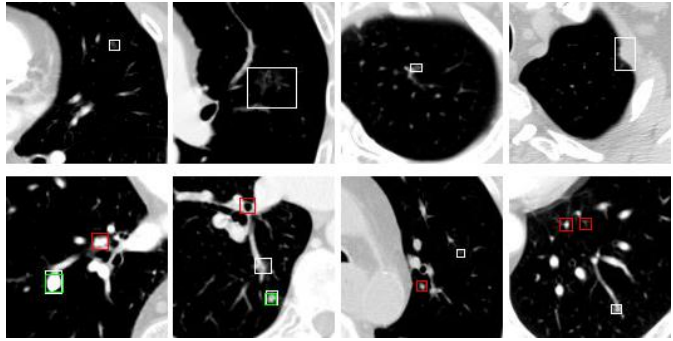

Figure 8. Examples of missing results (top row) and false-detected results (bottom row) by PiaNet on the LIDC-IDRI dataset, where the white rectangles denote the ground-truth boxes; the green rectangles denote the right results; and the red rectangles denote the wrong results.

# 6. Conclusions

In this paper, we have proposed a new convolutional neural network for detecting GGO nodules in 3D CT scans, which is named PiaNet. PiaNet is designed based on the idea of multiscale processing for addressing the problem of various sizes of GGOs. PiaNet is composed of a feature-extraction module, including a contracting pathway with pyramid multiscale source connections, an expanding pathway with skip connections, and a prediction module with multiscale outputs. Through the introduction of source connections and the integrated adoption of three multiscale processing strategies, PiaNet achieves satisfactory performance. It substantially and stably outperforms the state-of-the-art counterparts, including two methods, S4ND and GA-SSD, and two systems, SubsolidCAD and Aidence. The improvements in the CPM values compared with those obtained with these found conterparts are 3.26%, 1.85%, 20.25% and 15.07%, respectively. The ablation studies further justify the effectiveness of each of the three multiscale processing strategies. From these experimental results, we believe that PiaNet offers a promising tool for GGO nodule detection in the clinical diagnosis of lung cancer.

Although PiaNet can detect the majority of nodules with different sizes and irregular shapes and textures, it still misses some very uncommon nodules and some nodules with very low contrast and incurs false-detected nodules that are located very close to other tissues with similar appearances. It may be helpful to further reduce the error rate by making the network training focus on these hard examples. We plan to explore this approach in future studies.

**Funding**: This work was supported in part by the National Natural Science Foundation of China [grant numbers 60973059, 81171407, 61901533]; the Program for New Century Excellent Talents in University of China [grant number NCET-10-0044]; the Beijing Municipal Science and Technology Project [grant number Z181100001918002].

**Declaration of competing interest**: All co-authors of this manuscript confirm that there are no financial or personal relationships with any people or organizations that could inappropriately influence the actions of any author of this manuscript.